\documentclass[letterpaper]{article} 
\usepackage{aaai25}  
\usepackage{times}  
\usepackage{helvet}  
\usepackage{courier}  
\usepackage[hyphens]{url}  
\usepackage{graphicx} 
\usepackage{natbib}  
\usepackage{caption} 
\usepackage{algorithm}
\usepackage{algorithmic}

\usepackage{xcolor}
\usepackage{times}
\usepackage{latexsym}
\usepackage{graphicx}
\usepackage{booktabs}
\usepackage{multirow}
\usepackage{amssymb}
\usepackage{float}
\usepackage{ragged2e}
\usepackage{tabularx}

\usepackage{cleveref}
\graphicspath{ {images/} }
\usepackage{color,soul}
\usepackage[T1]{fontenc}

\crefformat{section}{\S#2#1#3} 

\usepackage{pifont}
\newcommand{\cmark}{\ding{51}}%
\newcommand{\xmark}{\ding{55}}%

\usepackage{inconsolata}
\usepackage{enumitem}

\usepackage{amsthm}

\theoremstyle{definition}
\newtheorem{definition}{Definition}

\usepackage[first=0,last=9]{lcg}

\newcommand\tab[1][0.5cm]{\hspace*{#1}}
\newcommand\smalltab[1][2mm]{\hspace*{#1}}
\usepackage{color, colortbl}
\definecolor{Gray}{gray}{0.9}
\definecolor{LightCyan}{rgb}{0.88,1,1}

\definecolor{lightsalmon}{RGB}{255, 230, 230}
\definecolor{darksalmon}{RGB}{255, 194, 194}
\definecolor{darkgreen}{RGB}{0, 100, 0}
\definecolor{mygreen}{RGB}{165, 243, 165}
\definecolor{palegreen}{RGB}{217, 246, 217}
\definecolor{palestgreen}{RGB}{235, 250, 235}
\definecolor{lightblue}{RGB}{204, 243, 255}
\definecolor{theblue}{RGB}{201, 218, 248}
\definecolor{thepurple}{RGB}{218, 202, 255}
\definecolor{vermilion}{RGB}{255, 60, 0}
\definecolor{lightgray}{RGB}{219, 219, 219}
\definecolor{mypurple}{RGB}{101, 25, 189}

\usepackage{newfloat}
\usepackage{listings}
\DeclareCaptionStyle{ruled}{labelfont=normalfont,labelsep=colon,strut=off} 
\floatstyle{ruled}
\newfloat{listing}{tb}{lst}{}
\floatname{listing}{Listing}
\title{Chain-of-Instructions: Compositional Instruction Tuning on \\
Large Language Models}
\author{
    Shirley Anugrah Hayati\thanks{\smalltab Work was partially done during internship at Amazon.}\smalltab \raisebox{3pt}{{\includegraphics[height=1em,width=1em]{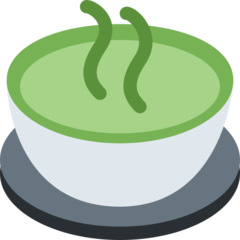}}} \tab Taehee Jung\raisebox{3pt}{{\includegraphics[height=1em,width=1em]{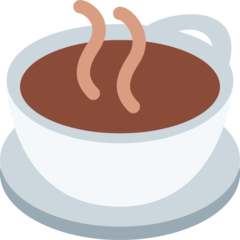}}} \tab Tristan Bodding-Long\raisebox{3pt}{{\includegraphics[height=1em,width=1em]{emoji/coffee.png}}} \\
    \tab Sudipta Kar\raisebox{3pt}{{\includegraphics[height=1em,width=1em]{emoji/coffee.png}}} \tab Abhinav Sethy\thanks{\smalltab Work was done while at Amazon.}\raisebox{3pt}{{\includegraphics[height=1em,width=1em]{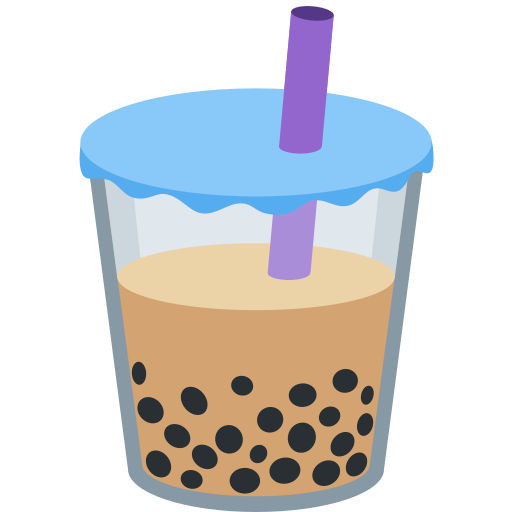}}} \tab Joo-Kyung Kim\raisebox{3pt}{{\includegraphics[height=1em,width=1em]{emoji/coffee.png}}}  \tab Dongyeop Kang\raisebox{3pt}{{\includegraphics[height=1em,width=1em]{emoji/tea.png}}}\\
}
\affiliations{
     \raisebox{3pt}{{\includegraphics[height=1em,width=1em]{emoji/tea.png}}}University of Minnesota \tab \raisebox{3pt}{{\includegraphics[height=1em,width=1em]{emoji/coffee.png}}}Amazon \tab \raisebox{3pt}{{\includegraphics[height=1em,width=1em]{emoji/bubbletea.png}}}Grammarly \\
    \{hayat023, dongyeop\}@umn.edu \smalltab \{jungtaeh, boddint, sudipkar, jookyk\}@amazon.com \smalltab
    abhinav.sethy@grammarly.com
}

\usepackage{bibentry}

\begin{document}

\maketitle

\begin{abstract}
Fine-tuning large language models (LLMs) with a collection of large and diverse instructions has improved the model's generalization to different tasks, even for unseen tasks. However, most existing instruction datasets include only single instructions, and they struggle to follow complex instructions composed of multiple subtasks.
In this work, we propose a novel concept of compositional instructions called \textit{chain-of-instructions} (CoI), where the output of one instruction becomes an input for the next like a chain. 
Unlike the conventional practice of solving single instruction tasks, our proposed method encourages a model to solve each subtask step by step until the final answer is reached.
CoI-tuning (i.e., fine-tuning with CoI instructions) improves the model's ability to handle instructions composed of multiple subtasks as well as unseen composite tasks such as multilingual summarization.
Overall, our study finds that simple CoI tuning of existing instruction data can provide consistent generalization to solve more complex, unseen, and longer chains of instructions. Our code and data are available at \url{https://github.com/amazon-science/chain-of-instructions}.

\end{abstract}

%

\section{Introduction}
Large language models (LLMs) have demonstrated impressive performance in various tasks, from conventional NLP downstream tasks, such as machine translation and summarization, to open-ended tasks, such as writing an outline for blog posts and giving tips for presentation, when fine-tuned on human-like instructions \cite{ouyang2022training, wang-etal-2022-super, DatabricksBlog2023DollyV2, mishra-etal-2022-cross}. 
These models excel at single instruction tasks, but their ability to handle complex and compositional instructions is less explored. 

\begin{figure}
    \centering
    \includegraphics[width=0.9\columnwidth, trim={0cm 0cm 0cm 0cm},clip]{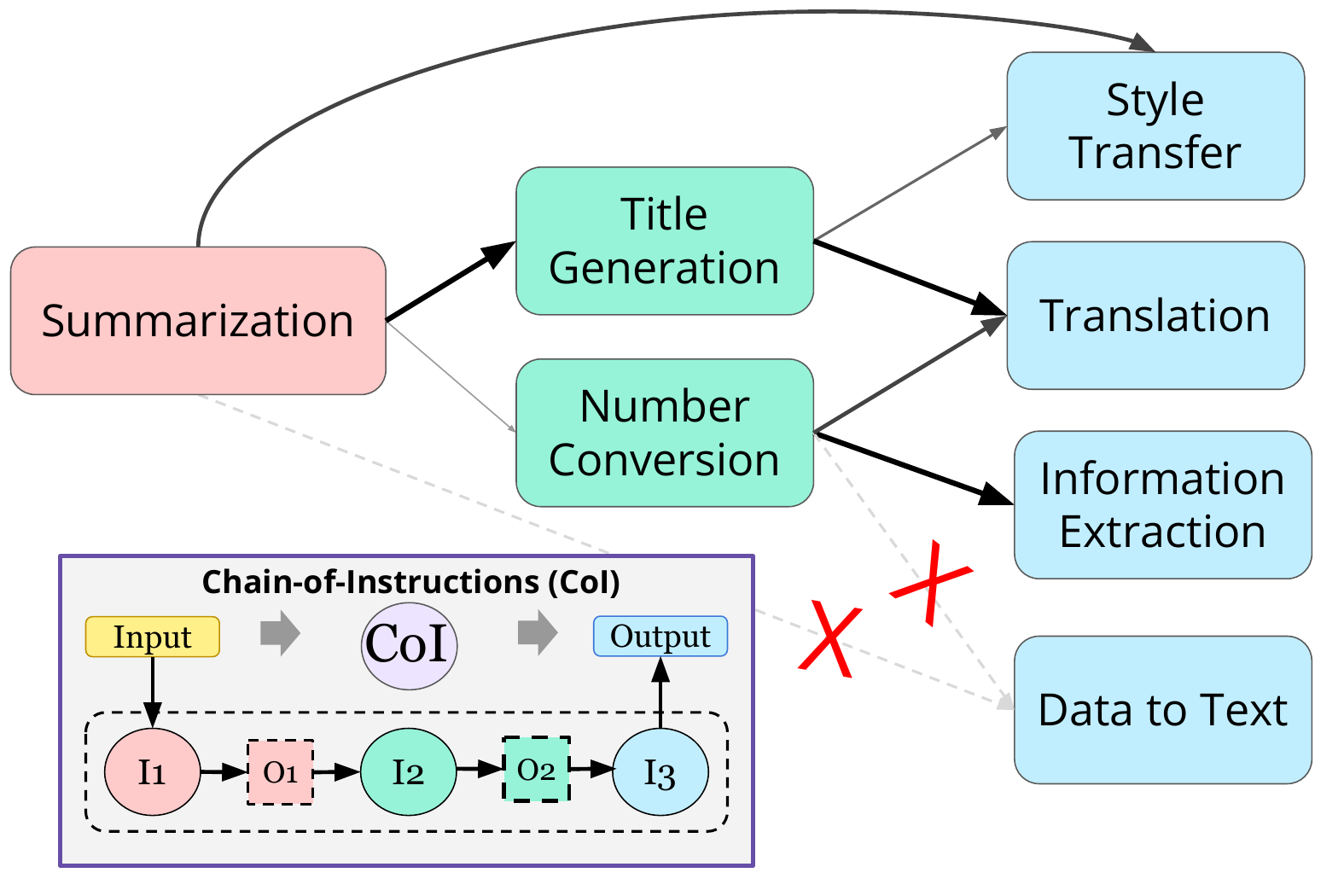}
    \caption{
    Chain-of-Instructions (CoI) example. The summarization output can be an input for a title generation subtask; the output of the title generation can be an input for style transfer or translation subtasks. 
    Arrow thickness denotes the probability of instruction composability. 
    \textcolor{red}{X} means that these subtasks cannot be composed due to format mismatch. 
    $I_{k}$ is $k^{th}$ instruction and $O_{k}$ is $k^{th}$ output.
    \vspace{-5mm}
    }
    \label{fig:figure1}
\end{figure}

A compositional instruction contains a series of sequential subtasks, as the output of one subtask becomes the input of the next one in a chained manner as shown in Figure \ref{fig:figure1}. We call this problem as \textit{Chain-of-Instructions} or shortly CoI. We examine what subtasks can be composed more naturally than others.

\begin{figure}
    \centering
    \includegraphics[width=\columnwidth, trim={0cm 0.9cm 0cm 19.5cm},clip]{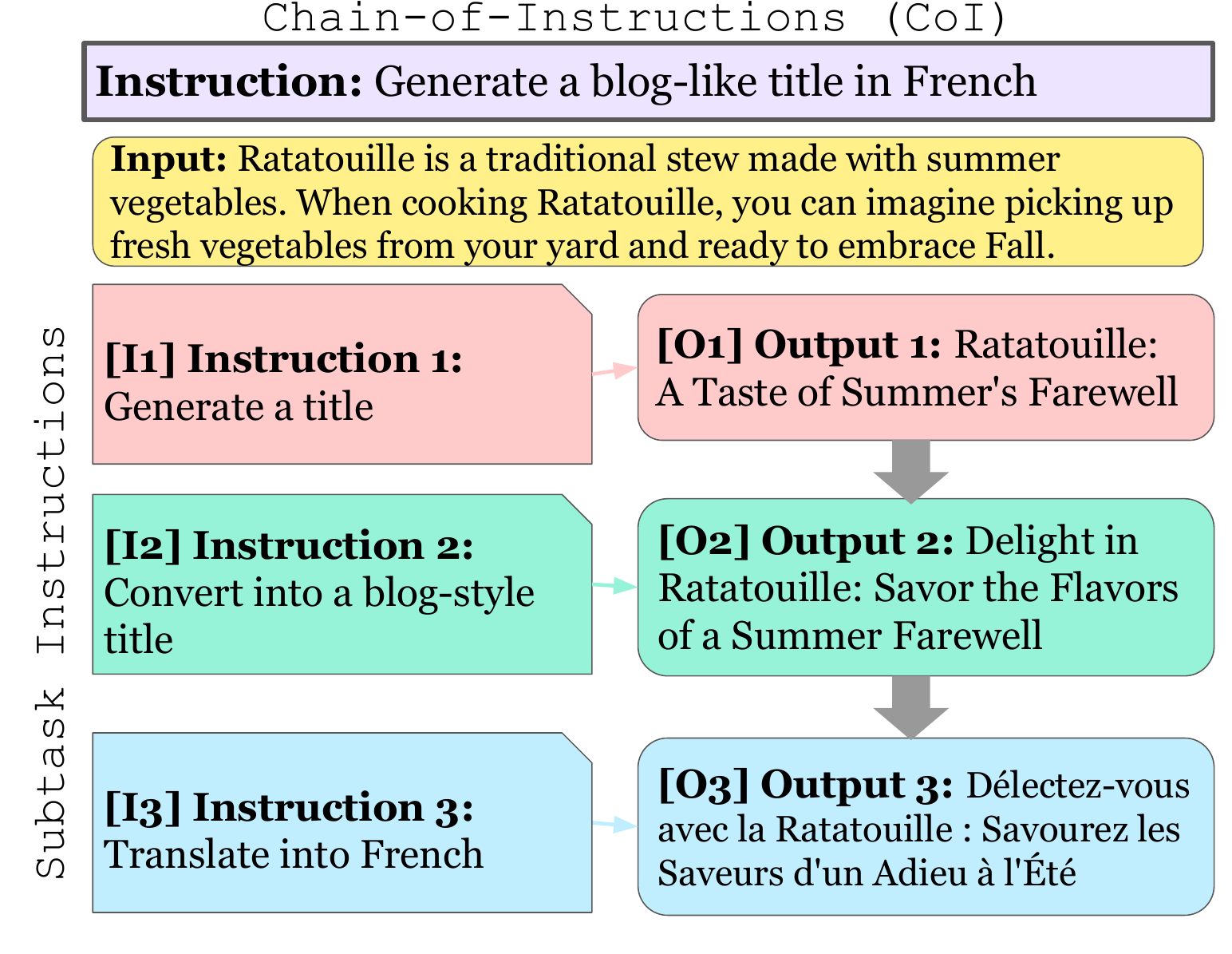}
    \caption{An example of the Chain-of-Instructions task. 
    The \colorbox{cyan!30}{last output} is the expected output of the \colorbox{violet!20}{CoI}. 
    \vspace{-3mm}
    }
    \label{fig:task_example}
\end{figure}

In Figure \ref{fig:figure1}, we can see that some tasks can be composed together, such as input a summary to a title generation task, while some tasks cannot be composed together, e.g., a summary as input for a Data-to-Text task. 
Some tasks have a higher probability of being able to be composed, such as generating a title from a summary compared to converting numbers in a summary since sometimes a summary does not contain a number. Figure \ref{fig:task_example} illustrates a more detailed example of CoI. The given instruction ``\textit{Generate a blog-like title in French}'' can be decomposed into three chained sub-instructions: 
\begin{enumerate}[noitemsep,topsep=2pt,leftmargin=*]
    \item Generate a title for the given text
    \item Convert the style of the title to be similar to a blog post title
    \item Translate the blog post title into French
\end{enumerate}

\begin{table*}[ht!]
\centering
\begin{tabular}{lcccc}
\toprule
 & Instruction & Composed & Data Size & Domain\\
\midrule
\textbf{Chain-of-Instructions} (Ours) &  \cmark & \cmark & 18k & NLP tasks\\ 
\midrule
Self-Instruct \cite{wang-etal-2023-self-instruct} &  \cmark & \xmark & 52k & Daily QA-like tasks\\
Dolly \cite{DatabricksBlog2023DollyV2}  & \cmark & \xmark & 15k  & Daily QA-like tasks\\
Super-NaturalInstruct \cite{wang-etal-2022-super}   & \cmark & \xmark & 1.6k  & NLP tasks\\\midrule
Faith and Fate \cite{dziri2023faith} & \xmark & \cmark & N/A  & Math, logic, programming\\
Compositional Semantic \cite{drozdov2022compositional} & \xmark & \cmark & N/A  & CFQ, COGS, Parsing\\
MuSiQue \cite{trivedi-etal-2022-musique-new}  & \xmark & \cmark & 24.8k & Multi-hop QA\\
\bottomrule
\end{tabular}
\caption{A comparison of our work with existing related works. As some previous works do not contribute a new dataset, the dataset size is shown as N/A. For instruction datasets, data size refers to the number of instructions, not task instances (input-output pair). 
More prior work is studied in \S\ref{sec:related}.
\label{tab:comparison_table} }
\end{table*}

When these sub-instructions are composed, we call it a compositional instruction or chain-of-instructions. Our study investigates whether LLMs can handle compositional instructions effectively and whether models tuned with compositional instructions can be generalized to solve more \textit{complex}, \textit{unseen}, or \textit{longer chains of instructions}. 
We first create a new CoI dataset with our proposed LLM-based compositionality checker, and then evaluate our model's performance in handling (1) traditional single instructions and (2) compositional instructions.

Our work is closely related to other instruction-tuning works and compositional studies in NLP, as summarized in Table \ref{tab:comparison_table}.
\citet{wang-etal-2023-self-instruct,DatabricksBlog2023DollyV2, wang-etal-2022-super} propose new instruction datasets, but they only handle single instruction problems. 
Although our approach draws inspiration from Chain-of-Thought (CoT) prompting \cite{wei2022chain} or Least-to-Most prompting \cite{zhou2022least}, our CoI is not a prompting technique but a collection of chained instructions validated by an LLM, showing generalization in solving complex and compositional problems.
Our contributions are as follows:
\begin{itemize}
\item We introduce a novel task called Chain-of-Instructions (CoI) to examine LLMs' capabilities in following compositional instructions by creating a new benchmark dataset.
\item We develop a framework to automatically construct composed instruction datasets with minimal human supervision. The framework leverages in-context learning on existing single-instruction datasets to create CoIs.
\item 
We propose a method for enabling LLMs to solve compositional tasks in an explainable way. 
As an example, a model can generate incremental outputs at each step of a complex task chain. 
With CoI-tuning, step-by-step instruction following becomes easier, especially when dealing with instructions composed of multiple subtasks. 

\item
We demonstrate through experiments and analysis that the CoI-tuned model outperforms both individual instructions and sequential compositional instructions. 
By training on CoI data, the model achieves higher performance. This result also generalizes for unseen longer chain test sets and downstream tasks. 
\end{itemize}




\section{Chain-of-Instructions}

\subsection{Formulation}
Compositional instructions contain multiple subtask instructions where the output from one subtask becomes the input for the next subtask similarlity to a composition function in math. Thus, we formalize the problem of chain-of-instructions as follows:

\begin{definition}[Chain of Instructions]
Given a tuple of <instruction $I$, input $X$, output $Y$>, let $I(X) = Y$ refer that an LLM generates output $Y$ with instruction $I$ and input $X$.
A sequence of instructions $\{I_{1},...,I_{k}\}$ is a chain of instructions with length $k$ if $I_{i+1} \circ I_{i}(X_{i})=Y_{i+1}$, for all $i \in \{\mathbb{N}: 1 \leq i \leq k\}$. 


\end{definition}

\subsection{Automatic Dataset Creation Pipeline} 
\paragraph{Seed Datatsets}
We curate a new compositional instruction dataset from existing single task instruction dataset: \textsc{Super-NaturalInstructions} (\textsc{Sup-NatIns}) \cite{wang-etal-2022-super}. 
We select \textsc{Sup-NatIns} as the seed dataset because it contains a wide variety of tasks (1,616 unique tasks) from 76 categories, including text categorization, summarization, and machine translation. Each category contains many different NLP tasks. For example, under the text categorization category, there exist sarcasm detection and politeness classification tasks. 
Each task in \textsc{Sup-NatIns} contains human-written descriptions that can be considered as instructions and instances as pairs of inputs and outputs. 
We only select tasks with English (1,341 unique tasks) as their input language to make sure that the chain is connected. 
For our single-task instruction tuning data (CoI$_1$), we randomly sample 10 input-output pairs, resulting in 13,410 instances.

\begin{figure*}
    \centering
    \includegraphics[width=0.95\linewidth, trim={1.2cm 0.1cm 0.9cm 0cm},clip]{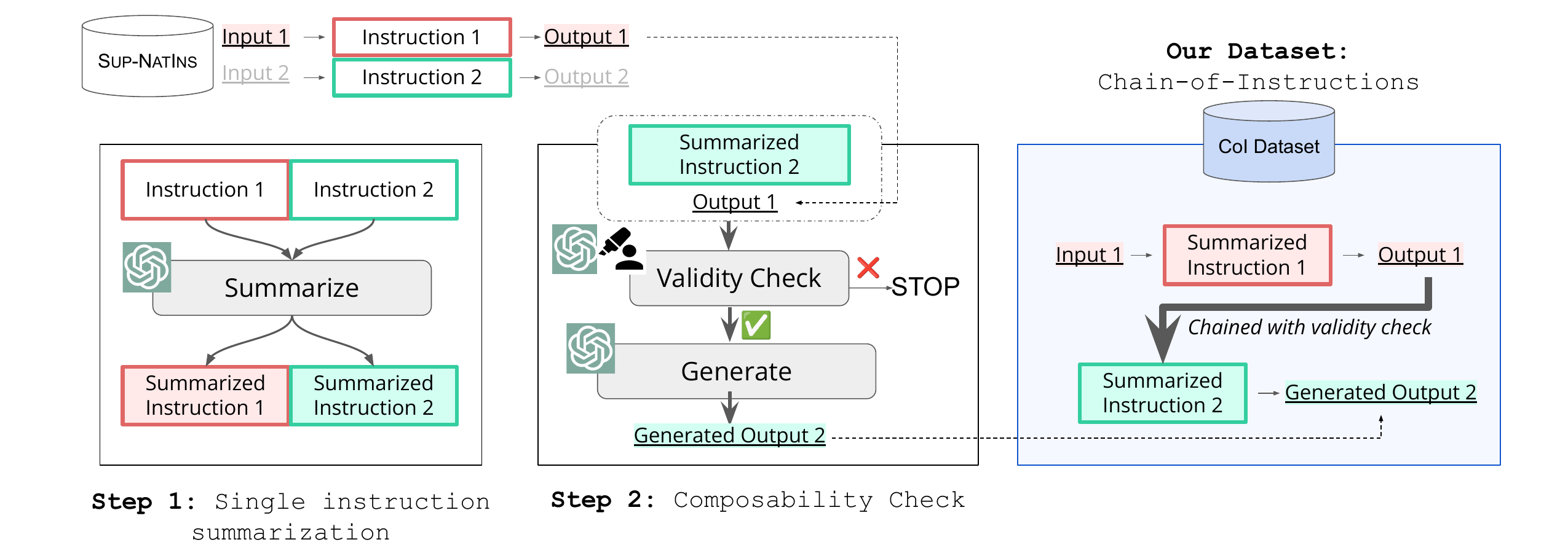}
    \caption{ Data creation for CoI$_{2}$. We use an LLM for both instruction summarization and composability check. The right column shows an example instance of our chain-of-instruction dataset. Output 1 in Step 2 comes from the original \textsc{SupNatInst} data. 
    }
    \label{fig:data_creation}
\end{figure*}

\paragraph{Instruction Composition}
Composing two single instructions poses a challenge due to their lengthy and specific descriptions, and differing output formats. Figure \ref{fig:data_creation} illustrates a two-step process for creating a compositional instruction dataset with the help of an LLM as elaborated in the following paragraphs. Here we use GPT 3.5 Turbo \cite{ouyang2022training} because of its reasonable price and at the time, we examine that the quality of the result is good enough. However, this data creation procedure can be reproducible with other strong LLMs as well. 

\paragraph{Step 1: Single instruction summarization}
The task instructions in \textsc{Sup-NatIns} are lengthy and detailed, which may deviate from real human-like instructions. With the same dataset (\textsc{Sup-NatIns}), \citet{yin-etal-2023-read} find that 60\% tokens can be removed with comparable, if not better, model performance. Thus, we use the LLM to shorten each instruction in the \textsc{Sup-NatIns} dataset. This step reduces the average number of words in the \textsc{Sup-NatIns} descriptions from 62.19 to 14.33. 

\paragraph{Step 2: Composability check}
To generate compositional instructions from single instructions, we perform a two-step process: (1) validity check and (2) generate the output for the second (or third) subtask. 
The validity check is performed to examine whether two subtasks are composable. We first filter out non-composable tasks with heuristics developed by the authors' knowledge (the Heuristics for Validity Check section in the Appendix). For example, classification tasks can only be the last subtask when composing a pair of tasks. After applying these heuristics, we additionally check whether LLM can generate the output for the second instruction based on the input of the first instruction. If so, we treat the pair as composable.\footnote{Prompt for this step is available in the Appendix.} 

For the pairs that pass the validity check, we generate the new output using the first output and second instruction for the second task.
This generated output serves as the ground truth for the second subtask in the instruction-tuning phase. Our approach is a variation of distillation from a larger LLM as has been done by previous works for different problems \cite{gu2024knowledge, hsieh-etal-2023-distilling, west-etal-2022-symbolic}.
We define compositional instructions originating from two instructions as CoI$_2$ and those originating from three instructions CoI$_3$. 
CoI$_3$ is created by chaining two CoI$_2$s if there exists I$_x \circ$  I$_y$ and I$_y \circ$ I$_z$, resulting in CoI$_3$ = I$_x \circ$  I$_y \circ$ I$_z$. The same method is applied for creating longer chains such as CoI$_4$ and CoI$_5$.

\begin{table}[t]
    \centering
    \begin{tabular}{c c c}
    \toprule
chain length ($\sigma$) & train & test
\\ \midrule
    1 & 13,410 & - \\ 2 & 2,993 & 588\\
    3 & 2,187 & 480 \\
    4 & - & 844\\
    5 & - & 355\\
    \bottomrule
    \end{tabular}
    \caption{Dataset statistics per chain length.}
    \label{tab:data_stats_main}
\end{table}

To examine the quality of LLM's composability check, we randomly sampled 100 instances and manually inspected which composed instructions are valid. We find that 75\% are valid composed instructions. For CoI$_3$, similarly we randomly sampled 100 instances and found that 59\% are valid compositions. Such error rates are often found in LLM-generated data \cite{das2024under, wang-etal-2023-self-instruct}.



\subsection{CoI Dataset}
Table~\ref{tab:data_stats_main} shows the data statistics of CoI datasets.
In chain length 2, we obtain 970 unique category pairs; in chain length 3, we obtain 418 unique category triplets. 
In each pair or triplet, we randomly select at most three instances and divide them into training and testing sets. 
For the longer chains (4, 5), we only use them for testing. 
Please find Appendix~\ref{subsec:data_stats} for the detailed statistics.

\begin{figure}
    \centering
    \includegraphics[width=1.03\columnwidth, trim={0.5cm 2.8cm 0cm 0.2cm},clip]{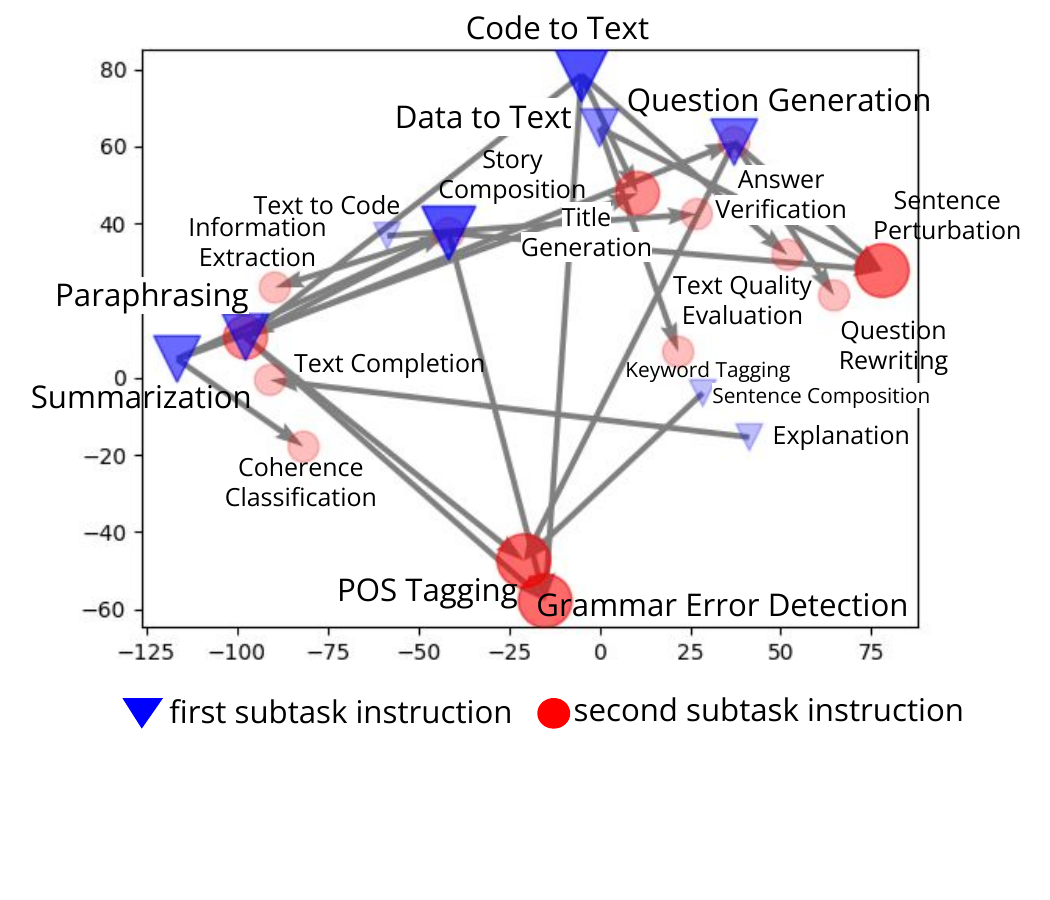}
    \caption{T-SNE of sentence embeddings for most frequent compositional instructions with CoI$_2$.
    }
    \label{fig:tsne_coi2}
\end{figure}

Figure \ref{fig:tsne_coi2} shows a t-SNE plot when we embed  subtask instructions of frequent CoI$_2$ instructions using SentenceBERT \cite{reimers-gurevych-2019-sentence} with DistilRoberta \cite{sanh2019distilbert}.\footnote{We only select instruction pairs that appear more than 7 times, and 9 is a max number of occurrences in CoI$_2$ dataset.} 
We find generation tasks such as paraphrasing and question generation can be compiled as both the first and second subtasks, except for problems involving specific input formats, such as code to text or data to text, which can only be compiled as the second subtask.
On the other hand, close-ended problems (e.g., POS tagging or grammar error detection) mostly appear as the second subtask. 

\section{Experiment Setup}

\paragraph{CoI models}
We fine-tune the base models of Alpaca-7B \cite{alpaca} and Mistral-7B-Instruct \cite{jiang2023mistral}. 
Since both models are open-sourced single instruction-tuned models which are widely used, they are suitable to be compared with CoI-tuned models. 

\paragraph{Baselines}
\begin{itemize}
    \item Off-the-shelf version of Alpaca-7B \cite{alpaca} model and Mistral-7B-Instruct model without fine-tuning (Base).
    \item The same non-finetuned Alpaca and Mistral with chain-of-thought prompting \cite{wei2022chain} (CoT) with seven-shot demonstrations and least-to-most prompting \cite{zhou2022least} (LtM). 
    \item Fine-tuned base models with a subset of single-instruction \textsc{Sup-NatIns} dataset (CoI$_1$). 
    
\end{itemize} 

\paragraph{Metrics}
For our evaluation metric, we report \textsc{Rouge-L} \cite{lin-2004-rouge},  following \citet{wang-etal-2022-super} and LLM (gpt-4o-mini) as a preference judge. \textsc{Rouge} can be used to assess various text generation tasks and using LLM as a judge has been widely adopted in NLP research \cite{liu-etal-2023-g,fu-etal-2024-gptscore}. We also have human evaluation to perform blind pairwise comparison between the outputs from the baseline and from our best CoI models. 


\paragraph{Test sets}

To assess the compositionality of our models, we prepare three types of evaluation suites.
\begin{itemize}
    \item \textbf{CoI Test set} For the compositional instruction evaluation, we tested the models on CoI test sets with $\sigma=\{2, 3, 4, 5\}$ where $\sigma$ is a chain length.
    \item \textbf{BIG-Bench Hard} For the single instruction test set, we use BIG-Bench Hard \cite{suzgun2022challenging}, a collection of 27 challenging tasks, such as date understanding and evaluating the truth value of Boolean expressions, and each task has $\le$ 250 instances. BIG-Bench Hard subset enables us to evaluate the model's performance on diverse and challenging NLP tasks with clear single instructions and associated input-output pairs. 
    \item \textbf{Downstream Task} In addition to CoI test sets, we examine the usefulness of CoI on the downstream task of multilingual summarization using WikiLingua \cite{ladhak-etal-2020-wikilingua}, which is a multilingual dataset based on WikiHow\footnote{\url{https://www.wikihow.com}} for abstractive summarization in 18 languages. WikiHow articles provide step-by-step instructions to complete procedural tasks on different topics, and each step includes a one-sentence summary as well. In our experiment, we select source-target language pairs 
    ; English-to-French (\textit{WikiLingua-en-fr}) and Spanish-to-English (\textit{WikiLingua-es-en}) and randomly sample 300 test instances for each. Given an input content from source language $L_{src}$, we aim to generate a summary in target language $L_{tgt}$. 
    This task is similar to a 2-instruction problem as we summarize first and then translate. Note that CoI training data only contains translation tasks from English to Punjabi, German, and Catalan, thus, selected source-target pairs are unseen in CoI training set.
\end{itemize}

\begin{table}[t!]
    \centering
    \begin{tabular}{l | r r | r r} 
    \toprule
      &   \multicolumn{2}{c|}{\textbf{Mistral}}  & \multicolumn{2}{c}{\textbf{Alpaca}} \\
      & Base & CoI & Base & CoI 
      \\ \midrule
     \multicolumn{5}{c}{\textbf{Test Set: CoI$_2$}}
     \\
     \midrule
        Subtask 1 &  1.32 & \textbf{90.50} &  13.98 & \textbf{84.16}
        \\
        Subtask 2 & 2.40 & \textbf{49.21} & 7.02 & \textbf{45.57}\\
    \midrule
     \multicolumn{5}{c}{\textbf{Test Set: CoI$_3$}}
     \\ \midrule
    Subtask 1 & 18.04 & \textbf{81.49} & 9.56 & \textbf{91.77}\\
    Subtask 2 & 6.82 & \textbf{68.65} &  2.13& \textbf{71.67}\\
    Subtask 3 & 6.93 & \textbf{32.73} &  3.30& \textbf{35.52}
        \\ 
    \bottomrule
    \end{tabular}
    \caption{\textsc{Rouge-L} 
 results on intermediate tasks. CoI models refer to best models of CoI: CoI$_{12}$ model if the test set=CoI$_2$ and CoI$_{123}$ model if the test set=CoI$_3$.
    }
    \label{tab:intermediate_result}
\end{table}

\section{Results}
We conduct experiments to measure the performance of CoI-tuned models on our compositional instructions (\S\ref{sec:coi_experiments}), and the generalization capability to difficult single instructions (\S\ref{sec:single_instruction_results}), 
and longer-chain instructions $\sigma = \{4,5\}$ (\S\ref{sec:coi45_instruction_results}),
and the application to an existing downstream task (\S\ref{sec:downstream_results}). We also conduct an ablation study to see if the correctness of second and third subtask outputs matter in the Table~\ref{tab:full_results} in Appendix. We see degrading performance of models fine-tuned with incorrect outputs, showing the importance to have the correct output for the subtasks during training.

\begin{table}[]
    \centering
    \small
    \addtolength{\tabcolsep}{-0.3em}
    \begin{tabular}{@{}l| r r | r r | r r@{}}
    \toprule
    \multirow{2}{*}[-1ex]{Model}& \multicolumn{2}{c|}{CoI$_2$-test}  & \multicolumn{2}{c|}{CoI$_3$-test}
  &  \multicolumn{2}{c}{BBH}\\  \cmidrule{2-7}
   & Mistral & Alpaca & Mistral & Alpaca & Mistral & Alpaca\\ \midrule
    \multicolumn{7}{l}{\textbf{Baselines}} \\
    \midrule
    Base &  24.93 &  24.95 & 23.66 & 20.99 & 8.51 & 14.36 \\
    CoT &  16.61 &23.82  & 16.90 &20.09 &  5.84 & 17.05\\
    LtM &  15.07 & 23.54 &  16.41 & 19.79 & 3.99 & 3.99 \\
    CoI$_1$ & 39.72 & 32.32  &29.62& 21.75 & 27.68 & 28.74 
    \\
    \midrule
    \multicolumn{7}{l}{\textbf{Chain-of-Instructions Models}} \\
    \midrule
    CoI$_2$ & 60.43 & 62.04 & 48.31 & 48.23 & 10.65 & 12.11 \\
    CoI$_3$  & 33.63 & 31.62 & 60.03 & 47.03 & 5.78 & 7.00\\
    CoI$_{12}$ & \colorbox{cyan!30}{\textbf{70.76}} & \colorbox{cyan!30}{\textbf{67.50}} &  59.84  & 50.23 &  24.44 & \colorbox{cyan!30}{\textbf{28.80}} \\
    CoI$_{123}$  & 45.16 & 67.12 & \colorbox{cyan!30}{\textbf{61.61}} &  \colorbox{cyan!30}{\textbf{67.49}} & \colorbox{cyan!30}{\textbf{29.39}} & 27.57 \\ 
    \bottomrule
    \end{tabular}
    \caption{\textsc{Rouge-L} results on compositional instruction test sets and BIG-Bench Hard (BBH). Base refers to the non-fine-tuned base models, CoT = chain-of-thought prompting on base models, LtM = least-to-most prompting on base models.
    The best scores are marked as \colorbox{cyan!30}{\textbf{bold}}.
    }
    \label{tab:multi_instruction_experiment}
\end{table}

\begin{table*}
\centering
    \small
    \begin{tabular}{p{0.8cm}|p{12cm}|p{1.8cm}}
    \toprule
    \multicolumn{3}{p{14.6cm}}{\textbf{Instruction: } \textcolor{red}{Simplify the given sentence by paraphrasing it.}}\\
    \multicolumn{3}{p{14.6cm}}{and then \textcolor{darkgreen}{Determine if the paraphrased sentence has proper punctuation with True or False.}}\\\midrule
     \multicolumn{3}{p{14.6cm}}{\textbf{Input:} {many of the churches work together for town-wide projects under the banner of "churches together in stevenage".}}\\\midrule
    \multicolumn{3}{p{14.6cm}}{\textbf{Correct output:} Task 1 output and task 2 input: \textcolor{red}{many of the churches work together for projects across the town under the slogan of ``churches together in stevenage''}. Task 2 output: \textcolor{darkgreen}{False}} \\
    \midrule
    \textbf{Model} & \textbf{Output} & \textbf{\textsc{Rouge-L}}\\ \midrule
        CoI$_1$ & many of the churches work together for town wide projects under the banner of `` churches & 75.00 \\
        &  together in stevenage ''.  &\\\midrule
        CoI$_2$ &Task 1 output and task 2 input: many of the churches work together for projects across the   & \textbf{100} \\
        & town  under the slogan of `` churches together in stevenage ''. Task 2 output: False & \\
        \midrule
        CoI$_3$ & 1 output and 2 input: many of the churches work together for projects across the town under the  & 43.75 \\
        &  slogan of `` churches together in stevenage ''.  2 output and 3 input: The pronoun 'them' refers to \\
        &  the noun phrase 'many of the churches' because the sentence states that 'they' work together for\\
        &   projects across the town.  This coreference is justified by the knowledge that the churches are the ones that are working together. 3 output: False \\
        \midrule
        CoI$_{12}$ & 1 output and 1 input: many of the churches work together for projects across the town under the & 85.71\\
        &slogan of `` churches together in stevenage ''. 2 output: False  & \\\midrule
        CoI$_{123}$  &Task 1 output and task 2 input: many of the churches work together for projects across the town   & \textbf{100} \\
        &  under the slogan of `` churches together in stevenage ''. Task 2 output: False & \\
        \bottomrule
    \end{tabular}
    \caption{Example outputs from different CoI-tuned Alpaca on a 2-instruction test instance. 
    }
    \label{tab:qual_multiple}
\end{table*}

\subsection{Performance on In-domain Composite Tasks (CoI$_{2,3}$)}
\label{sec:coi_experiments}
\paragraph{Automatic metric}
As we evaluate our CoI models' performance against the baselines on multi-CoI test sets, we find that both Mistral and Alpaca fine-tuned on CoI$_{12}$ instructions perform the best for CoI$_2$-test (Table \ref{tab:multi_instruction_experiment}) \footnote{Fine-tuning details in Appendix.}. Similarly, for CoI$_3$-test, both CoI$_{123}$ Mistral and Alpaca perform the best. All models fine-tuned on compositional instructions generally outperform the baselines, except for CoI$_3$-tuned Alpaca. This model performs slightly worse than the CoI$_1$-tuned Alpaca on CoI$_2$ test set. We hypothesize that this happens because instructions in CoI$_3$ become very long, 
thereby it becomes harder for the model to generalize without CoI$_2$ and CoI$_1$ examples. As a result, models only fine-tuned on CoI$_3$ tend to generate long sentences with hallucinations as in Table \ref{tab:qual_multiple}. 

In the LLM-as-a-judge experiment, we evaluate the performance of the best CoI models on CoI$_2$-test and CoI$_3$-test against the best baseline, CoI$_1$. On CoI$_2$-test, the LLM prefers 69.90\% of Alpaca CoI$_{12}$'s outputs and 70.92\% of Mistral CoI$_{12}$'s outputs over the baseline. Similarly, on CoI$_3$-test, the LLM favors Alpaca CoI$_{123}$'s outputs and Mistral CoI$_{123}$'s outputs over the baseline by 81.04\% and 60.00\%, respectively.

\paragraph{CoI Results per Subtasks}
We examine how CoI models perform for each subtask in the compositional instruction. To do this, we compare the results from the best version of CoI (CoI$_{12}$ for CoI$_{2}$ test set, CoI$_{123}$ for CoI$_{3}$ test set)
against the non-finetuned baseline models. 
Since there is no clear boundary to distinguish the first subtask output and the second subtask output in the baseline's outputs, we use an LLM to separate the responses.
Given the subtask instruction and the output, we ask the LLM to decide which span of the output text responds to the subtask instruction. To remove the possibility of LLM's hallucination being counted as part of the output, we only include LLM's output span when it appears in the baseline's output. When LLM deems that the output is incorrect, we assign \textsc{Rouge} = 0 because this output could refer to the first subtask or second subtask.

\begin{figure} 
    \centering
    \includegraphics[width=0.9\columnwidth, trim={2cm 2.3cm 2.3cm 2cm},clip]{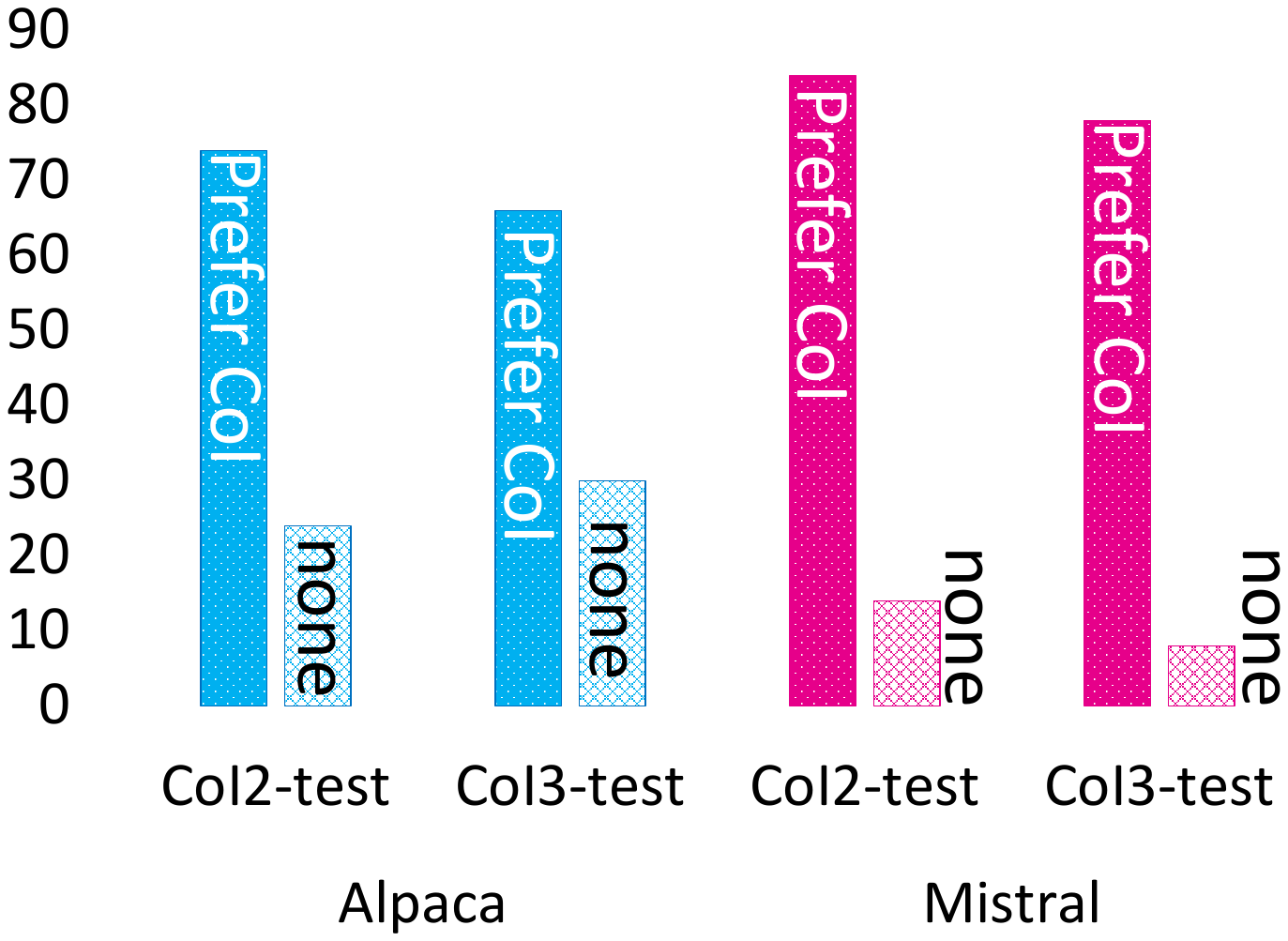 }
    \caption{Human evaluation results. "Prefer CoI" refers to the percentage of CoI outputs preferred by humans; ``none'' refers to when humans think the outputs for both first and second subtasks are incorrect. 
    }
    \label{fig:human_eval}
\end{figure}

Table \ref{tab:intermediate_result} and Table \ref{tab:intermediate_result_valid} show results of CoI models and baseline on CoI$_{2}$ and CoI$_{3}$ test sets. In general, CoI models outperform the baseline for both test sets, with the best results on the first subtask and then followed by the second subtask and the third subtask. However, it is interesting to see that CoI's performance on the second subtask for CoI$_3$ test set is higher than CoI's performance on the second subtask for  CoI$_2$. We conjecture that it happens because the previous subtasks can be easier to solve or knowing the subsequent subtask helps improving the performance.

\paragraph{Human evaluation}
We randomly sample 200 instances from CoI$_2$-test and CoI$_3$-test for both Mistral and Alpaca and ask 8 computer science students with NLP knowledge to rate the output. 
We show 25 sets of instruction, input, baseline output, and CoI output to each annotator. The order of the outputs is randomized. Each annotator then selects the output they prefer.
Figure \ref{fig:human_eval} depicts the percentage of CoI outputs that humans prefer vs. when none of the outputs are preferred. Outputs from CoI-tuned models are preferred for all test sets for both models.


\begin{figure}
    \centering
    \includegraphics [width=0.75\columnwidth, trim={2cm 2.3cm 2.4cm 2cm},clip]{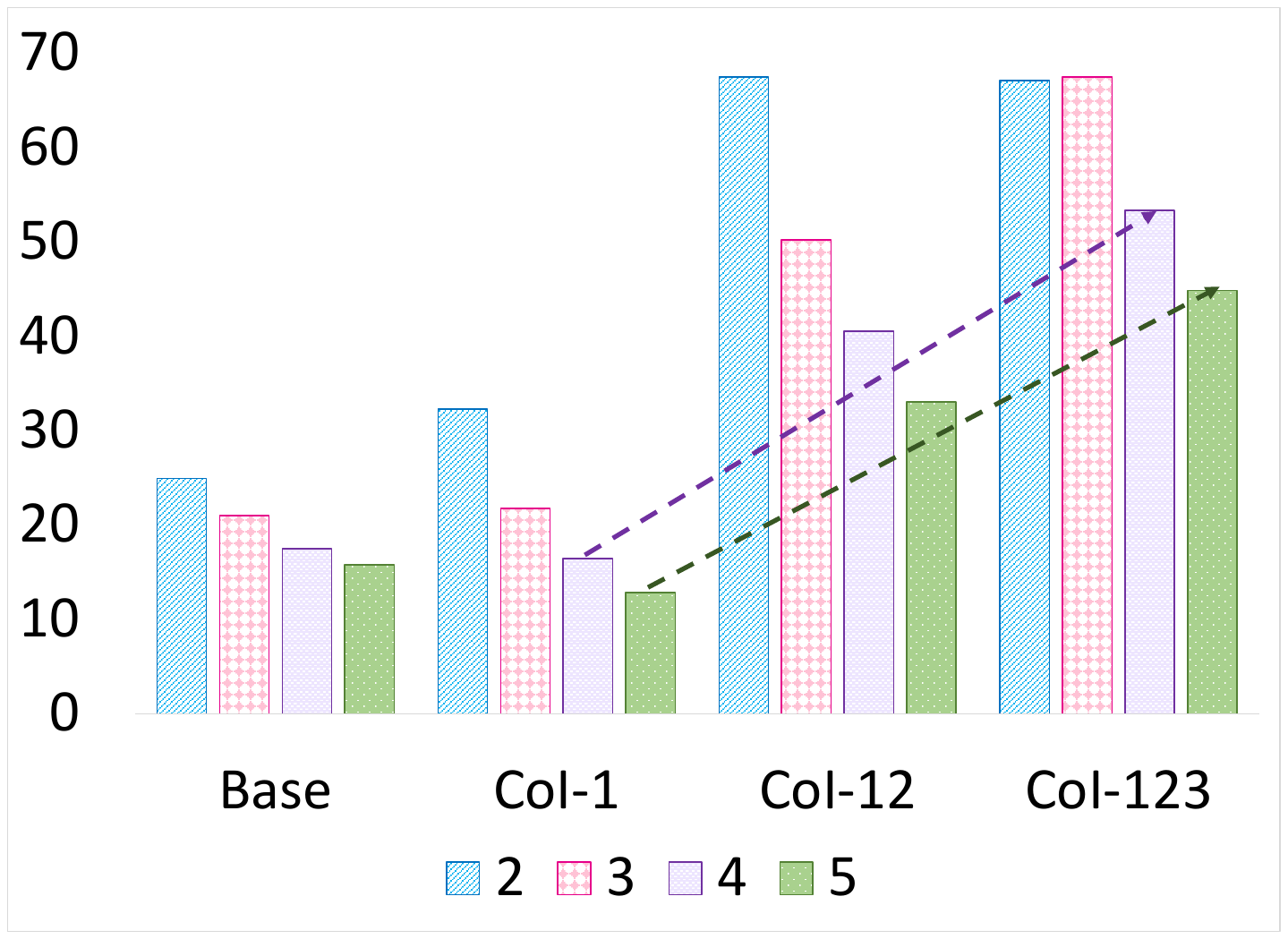}
    \caption{\textsc{Rouge}-L (x-axis) on CoI test sets $\sigma=2,3,4,5$ for various Alpaca models (y-axis). Base refers to the non-fine-tuned Alpaca.  
    }    
    \label{fig:coi_45}
\end{figure}

\subsection{Generalization to Unseen Single Tasks}
\label{sec:single_instruction_results}
To assess whether adding compositional instructions helps improve the model's performance on unseen and difficult single instruction tasks, we tested CoI-tuned models on BIG-Bench Hard (BBH). CoI$_{123}$-tuned Mistral performs the best (\textsc{Rouge}: 29.39) as shown in Table \ref{tab:multi_instruction_experiment}. For Alpaca, the model fine-tuned on CoI$_{12}$ is also better than the baseline and achieves a higher \textsc{Rouge} score of 28.80. This confirms that having compositional instructions helps the model to understand hard single instruction problems as well. 

\subsection{Generalization to Longer Chains (CoI$_{4,5}$)}
\label{sec:coi45_instruction_results}

In this experiment, we examine whether our CoI models can generalize to longer chains. We run inference on CoI$_{4}$ and CoI$_{5}$ test sets using CoI$_{1}$, CoI$_{12}$, and CoI$_{123}$-tuned Alpaca.\footnote{Mistral results are in Table \ref{tab:coi45test_experiment} in Appendix~\ref{appendix:fullresults}.} 
As shown in Figure \ref{fig:coi_45}, longer chains ($\sigma=2,3$) in the training set help the model to understand unseen longer chain ($\sigma=4,5$) in the test set as well. Moreover, the performance does not drop as high as CoI$_{1}$ or the baseline non-fine-tuned models that do not learn the chaining reasoning. We posit that the knowledge of compositional instructions in the training set, even though the length of the chain is shorter than 4 or 5, still helps the model to understand the composed tasks.


\subsection{Generalization to Downstream Composite Tasks}
\label{sec:downstream_results}
For this experiment, we use CoI$_{12}$ because it shows the highest \textsc{Rouge-L} on 2-instruction problem. For the baseline, we use non-finetuned Alpaca and Mistral. 
We evaluate the performance of the models using four metrics below.
\begin{itemize}
    \item \textbf{\textsc{Rouge-L} (all)} the \textsc{Rouge} score of the summary of the whole generated output. 
    \item \textbf{\textsc{Rouge-L} (src)} the \textsc{Rouge} score only from the summary in the source language.
    \item \textbf{\textsc{Rouge-L} (tgt)}  the \textsc{Rouge} score only from the summary in the target language.
    \item \textbf{\#valid outputs} number of valid summaries in the source 
    and the target languages 
    are generated because sometimes the model may not generate them properly.
\end{itemize}

Table \ref{tab:downstream} shows the results for our downstream task experiments. 
For the English-to-French summarization task, CoI$_{12}$ can generate more valid target outputs than the baselines. 
Moreover, CoI$_{12}$ obtains higher \textsc{Rouge} for both source and target summaries than the baselines. 
For the Spanish-to-English summarization task, CoI$_{12}$ Mistral outperforms the baseline for all \textsc{Rouge-L} scores, but Alpaca fails to have better \textsc{Rouge-L} (src) and \textsc{Rouge-L} (tgt) against the baseline.

In general, CoI performs better in English-to-French summarization compared to Spanish-to-English summarization because our training instances contain a translation task from English to other languages (Punjabi, German, and Catalan), even though the target language of the translation task in the training set is not French. 
On the other hand, we see poor performance in Spanish summaries across all models, possibly due to the lack of Spanish as the first subtask in training datasets.
We conjecture this issue could be resolved if we add more Spanish tasks during the fine-tuning stage.

\begin{table}[t]
    \centering
    \begin{tabular}{l | r r | r r} 
    \toprule
     \textbf{Metric} &   \multicolumn{2}{c|}{\textbf{Mistral}}  & \multicolumn{2}{c}{\textbf{Alpaca}} \\
      & Base & CoI$_{12}$ & Base & CoI$_{12}$ 
      \\ \midrule
     \multicolumn{5}{c}{\textbf{English to French}}
     \\
     \midrule
        \textsc{Rouge}-L (all)&  8.03  & \textbf{10.97}&5.78&\textbf{8.90} 
        \\
        \textsc{Rouge}-L (src)&  10.68 & \textbf{15.66} &3.84 & \textbf{12.71}
        \\
        \textsc{Rouge}-L (tgt) & 7.45  & \textbf{10.93}  & 5.46& \textbf{7.96} \\
        \hline
        \#valid src outputs  & 206  & \textbf{295} & 126 & \textbf{228}  \\
        \#valid tgt outputs  &  212  & \textbf{295} & 221 &  \textbf{228} 
        \\
    \midrule
     \multicolumn{5}{c}{\textbf{Spanish to English}}
     \\ \midrule
      \textsc{Rouge}-L (all) & 11.22  & \textbf{12.43} & 7.87 & \textbf{10.39} 
      \\
        \textsc{Rouge}-L (src) & 0.07 &  \textbf{4.85} & \textbf{2.47} &  1.87 
        \\
        \textsc{Rouge}-L (tgt) & 11.22 &\textbf{12.30} & \textbf{7.68} & 7.13  
        \\ 

        \hline
        \#valid src outputs  & 1  & \textbf{290} & 80  & \textbf{150}   \\
        \#valid tgt outputs & \textbf{300}  & 290 & \textbf{240} & 150 
        \\ \bottomrule
    \end{tabular}
    \caption{Results of the multilingual summarization task on 300 instances. Base refers to non-fine-tuned baseline, src is source language, and tgt is target language.}
    \label{tab:downstream}
\end{table}

\section{Related Work}\label{sec:related}

\paragraph{Instruction tuning}
There has been a notable surge in research focused on fine-tuning LLMs using human instructions. \citet{efrat2020turking} examined LLMs' ability to follow natural language instructions compared to crowd-workers. \citet{wei2022finetuned, sanh2021multitask} have transformed NLP task descriptions into human-like language instructions and showed that LLMs fine-tuned with those instructions have generalizable capability toward unseen tasks \cite{chung2024scaling}. Subsequently, many studies have emerged to create new instruction datasets aimed at training models in instruction-tuning paradigm: some instruction datasets are fully written by humans \cite{wang-etal-2022-super, DatabricksBlog2023DollyV2}, the others are written with the help of LLMs \cite{honovich-etal-2023-unnatural, wang-etal-2023-self-instruct, alpaca}; some instructions are NLP-specific \cite{ mishra-etal-2022-cross, wang-etal-2022-super, weller-etal-2020-learning}, and the others are designed to respond to general-purpose instructions \cite{ouyang2022training, wang-etal-2023-self-instruct}. 
These prior studies only work on single instruction datasets, so we construct a new compositional dataset upon \citet{wang-etal-2022-super}'s \textsc{Super-NaturalInstruction}. Our work is also related to several past works which have leveraged LLMs to generate training data \cite{schick-schutze-2021-generating}, and some of them specifically use LLMs for generating instruction data \cite{peng2023instruction, shao2023synthetic}. Nevertheless, our CoI data generation framework differs from previous works as we use LLMs to determine the composability of individual instructions, and then generate responses for subsequent subtask instructions. 


\paragraph{Compositional problems in NLP}
Several NLP work have investigated the capability of Transformer model on compositional problems including algorithm and math problems \cite{dziri2023faith}, compositional semantics \cite{drozdov2022compositional}, and multi-hop question-answering (QA) tasks \cite{trivedi-etal-2022-musique-new}. \citet{dziri2023faith} highlight how Transformer models often struggle with compositional mathematics computation or program executions \cite{nye2022show, saparov2022language}. \citet{drozdov2022compositional} introduce a new prompting method which first decomposes the compositional questions or sentences \cite{keysers2019measuring, kim-linzen-2020-cogs}, then sequentially predicts the answers to subproblems, and finally generating the final output. 
Compositionality in NLP is closely related with multi-hop QA problems with compositional questions where the answers from sub-questions are needed to answer the main question \cite{yang-etal-2018-hotpotqa, ho-etal-2020-constructing, trivedi-etal-2022-musique-new}. \citet{qiu-etal-2022-improving} have shown how a model with compositional latent structure improves large language models' performance on compositional generalization tasks through synthethic data augmentation. CoI is most related to \citet{aksu2023cesar} as they work on dealing with compositional tasks for dialogue systems. However, their definition of ``compositional task'' is different from ours as they do not require the output of one subtask is shown to the next subtask. Meanwhile, in our CoI, the outputs from the previous subtasks become the next subtask inputs. 


\section{Conclusion and Future Work}
In this work, we propose a new task called Chain-of-Instructions and develop a dataset for building models to solve the task. We introduce an automatic pipeline on how to build our dataset and demonstrate the usefulness of our CoI-tuned models on the tasks of the generated dataset and downstream tasks. Since human language is complex and an instruction may actually be composed of subtasks, it is important to have a model that can deal with compositional instructions, especially as we show that models fine-tuned only on single instructions never outperform the CoI-tuned models on multi-instruction tasks. For future work, we consider looking into instruction decomposition in addition to the instruction composition problem. We also recommend trying out more tasks to be composed besides those from \textsc{SuperNaturalInstruction}. 


\section*{Acknowledgements}
We would like to thank members of the MinnesotaNLP lab for their feedback and intellectual support. 


\bibliography{aaai25}

\section{Appendix}
\label{sec:appendix}
\subsection{CoI Dataset Statistics}
\label{subsec:data_stats}

\begin{table}[h]
\small
    \centering
    \begin{tabular}{l r | l r}
    \toprule
    \multicolumn{3}{l}{} & \textbf{Count}\\ \hline
    \multicolumn{3}{l}{Subtask 1 categories (CoI$_2$)} & 41\\
    \multicolumn{3}{l}{Subtask 2 categories (CoI$_2$)} & 67\\
    \multicolumn{3}{l}{\#Unique pairs of category 
    1$\rightarrow$2} & 970\\
    \multicolumn{3}{l}{Avg. \#words per instruction} & 34.25 \\
    \hline
    \multicolumn{3}{l}{Subtask 1 categories (CoI$_3$)}& 39\\
    \multicolumn{3}{l}{Subtask 2 categories (CoI$_3$)}& 35\\
    \multicolumn{3}{l}{Subtask 3 categories (CoI$_3$)}& 61\\ 
    \multicolumn{3}{l}{\#Unique pairs of category 1$\rightarrow$2} & 418\\
    \multicolumn{3}{l}{\#Unique pairs of category 2$\rightarrow$3} & 700\\
    \multicolumn{3}{l}{\#Unique triplets of category 1$\rightarrow$2$\rightarrow$3} & 2148 \\
    \multicolumn{3}{l}{Avg. \#words per instruction} & 51.00\\
    \midrule
    \multicolumn{2}{l|}{\textbf{Training Set Size}} & \multicolumn{2}{l}{\textbf{Test Set Size}}\\ \midrule
    CoI$_1$ & 13,410 & CoI$_2$ & 588\\
    CoI$_2$ & 2,933 & CoI$_3$ & 480\\
    CoI$_3$ & 2,187 & CoI$_4$ & 844\\
    CoI$_{12}$ & 16,343 & CoI$_5$ & 355\\
    CoI$_{123}$ & 18,530 & 
     \\
    \bottomrule
    \end{tabular}
    \caption{Dataset statistics}
    \vspace{-5mm}
    \label{tab:data_stats}
\end{table}

Table \ref{tab:data_stats} shows statistics for CoI. In CoI$_{2}$, the numbers of unique categories for the first subtask and second subtask are 41 and 67, respectively. There are more different categories for the last subtask because we set classification problems can only be the last subtask. There are 970 unique pairs of categories (e.g., summarization $\rightarrow$ title generation is counted as 1 pair). On average, a CoI$_{2}$ instruction contains 34.25 words. For CoI$_{3}$ instances, there are fewer categories to be the first subtask (39) and second subtask (35) because we need to guarantee the composability for the whole longer chain. 
When we compose the CoI datasets, we select at max 3 instances per pair of categories. We also report train and test set sizes in Table~\ref{tab:data_stats}.

\subsection{Prompting Details}
Here are prompting details for our approach.

\subsubsection{Summarizing and generalizing a single instruction}
\begin{itemize}[noitemsep,topsep=2pt,leftmargin=*]
    \item \textbf{Prompt}: ''Given an instruction and a category, simplify the instruction to have fewer than 30 words and make the instruction to be more general . 

Instruction 1: In this task, you're given passages that contain mentions of names of people, places, or things. Some of these mentions refer to the same person, place, or thing. Your job is to write questions that evaluate one's understanding of such references. Good questions are expected to link pronouns (she, her, him, his, their, etc.) or other mentions to people, places, or things to which they may refer. Do not ask questions that can be answered correctly without understanding the paragraph or having multiple answers. Avoid questions that do not link phrases referring to the same entity. For each of your questions, the answer should be one or more phrases in the paragraph, and it should be unambiguous.

Category 1: Question generation

Modified instruction 1: Generate a question given a paragraph that mentions people, places, or things''
    \item \textbf{Number of few-shot demonstrations}: 5
\end{itemize}

\subsubsection{Checking compositionality of two instructions and obtaining the second output}\label{appendix:check_composibility_second_output}
\begin{itemize}[noitemsep,topsep=2pt,leftmargin=*]
    \item \textbf{Prompt}: Decide whether the input is a valid input for the instruction. An input is valid if the context is relevant to instruction and we can generate an output given the input and the instruction. If it is a valid input, generate the output. 

Instruction: Categorize each of the following instruments as either string or keyboard: Guitar, Violin, piano, harmonium, cello, accordion, banjo
Input: The episode focused on two people: an elderly hospital doctor (Aleksander Bardini), who lived by himself in the ubiquitous Dekalog apartment block; and Dorota Geller (Krystyna Janda), a woman in her 30s who lived in the same building and was a violinist with the Philharmonic Orchestra.

Answer:

{"Valid input":"No", "Reason": "The instruction asks for categorizing instrument types and the instruction already contains its input which are guitar, violin, piano, harmonium, cello, accordion, and banjo. Meanwhile the given input is about a movie episode so it is not relevant to the instruction.", "Output": ""}

    \item \textbf{Number of few-shot demonstrations}: 5
\end{itemize}

\subsubsection{Making the input type of the second instruction consistent with the output type of the first instruction}\label{appendix:make_second_output_consistent}
\begin{itemize}[noitemsep,topsep=2pt,leftmargin=*]
    \item \textbf{Prompt}: You are given an instruction which is composed of multiple tasks. Modify the following instruction by first, identify the output1, input2, output2, input3, and so on related to the subtasks. Then make sure that output1 is consistent with input2, output2 is consistent with input3 and so on.

Instruction: "Who is the author of the Little Women and then how many capital letters are in the input?"

Subtask 1: "Who is the author of the Little Women?"

Subtask 2: "How many capital letters are in the input?"

Answer: \{"output1:" "author’s name", "input2:": "input", "modified\_instruction":  "Who is the author of the Little Women and then how many capital letters are in the author’s name?"\}

    \item \textbf{Number of few-shot demonstrations}: 8
\end{itemize}

\subsubsection{Separating outputs for the subtasks from Baseline Outputs}
\begin{itemize}
    \item \textbf{Prompt}: You are given an instruction, an input, and a text output. Decide which part of the text output responds to the given instruction for the given input. Don't change the wording of the text output! The text output may not correctly answer the instruction. If that's the case, you must respond with "Wrong"!

Instruction: \texttt{instruction}

Input: : \texttt{input}

Text Output: : \texttt{model-generated output}

Part of the text responding to the instruction: 

\item \textbf{Prompt}: You are given an instruction and a text. Decide which part of the given text responds to the given instruction. Don't change the wording of the given text! The text may not correctly answer the instruction. If that's the case, you must respond with "Wrong"!

Instruction: \texttt{instruction}
Text: \texttt{model-generated output}

Part of the text responding to the instruction:
\end{itemize}

\subsubsection{Creating concise-CoI}
\begin{itemize}[noitemsep,topsep=2pt,leftmargin=*]
    \item \textbf{Prompt}: Summarize the input to a single coherent sentence which contains only 20 words or fewer without changing the meaning.

Input: "Summarize the article in one sentence. and then Convert the sentence to a positive sentence with minimal changes. and then Extract RDF triplets from the given sentence."

Summary: "Create a concise summary with a positive tone and after that extract RDF triplets from the summary"
    \item \textbf{Number of few-shot demonstrations}: 2
\end{itemize}

\subsubsection{LLM as a Preference Judge}
The following is the prompt for LLM as a preference judge:
\newline
\newline
\noindent\fbox{\begin{minipage}{\columnwidth}
Given the following instruction and input and ground truth output, which generated output follows the instruction more closely? 
\newline
\newline
Answer only with ``A ''if you prefer generated output A, ``B'' if you prefer generated output B, ``None'' if you prefer none.
\newline
\newline
Instruction: \texttt{instruction}
Input: \texttt{input}
Ground truth output: \texttt{ground truth output}

Generated output A: \texttt{output from model A}
Generated output B: \texttt{output from model B}
\end{minipage}}
\subsection{Fine-tuning Details}
\label{fine-tuning_details}
Our base models to be fine-tuned are Alpaca-7B and Mistral-7B-Instruct. We use the training script and set-ups from \citet{alpaca}'s code. We fine-tune the whole model for 3 epochs with deepspeed on a machine with 8 NVIDIA A100 GPUs, batch size=4, learning rate = 2e-5, weight decay=0, warmup ratio = 0.03, max length = 512. Inference results are from single runs. 

\subsection{BIG-Bench Hard Examples}
\label{bbh_examples}
Here are some example tasks from BBH along with their instructions, example input, and example output:
\begin{enumerate}[noitemsep,topsep=2pt,leftmargin=*]
    \item \textbf{Boolean Expressions.} 
    
    Instruction: \texttt{Evaluate the truth value of a random Boolean expression consisting of Boolean constants (True, False) and basic Boolean operators (and, or, and not).}

Input: \texttt{not ( True ) and ( True ) is}

Output: \texttt{False}

    \item \textbf{Causal Judgment.}

    Instruction: \texttt{Given a short story (involving moral, intentional, or counterfactual analysis), determine how a typical person would answer a causal question about the story.}

    Input: \texttt{How would a typical person answer each of the following questions about causation?
    A machine is set up in such a way that it will short circuit if both the black wire and the red wire touch the battery at the same time. The machine will not short circuit if just one of these wires touches the battery. The black wire is designated as the one that is supposed to touch the battery, while the red wire is supposed to remain in some other part of the machine. One day, the black wire and the red wire both end up touching the battery at the same time. There is a short circuit. Did the black wire cause the short circuit? Options:- Yes - No}

    Output: \texttt{No}

    \item \textbf{Date Understanding.}
    
    Instruction: \texttt{Given a small set of sentences about a particular date, answer the provided question (e.g., ``The concert was scheduled to be on 06/01/1943, but was delayed by one day to today. What is the date yesterday in MM/DD/YYYY?'').}

    Input: \texttt{Today is Christmas Eve of 1937. What is the date tomorrow in MM/DD/YYYY?}
    
    \texttt{Options:}
    
    \texttt{(A) 12/11/1937} \\
    \texttt{(B) 12/25/1937} \\
    \texttt{(C) 01/04/1938}\\
    \texttt{(D) 12/04/1937} \\
    \texttt{(E) 12/25/2006} \\
    \texttt{(F) 07/25/1937}\\

    Output: \texttt{(B)}
    \item \textbf{Disambiguation QA.} 
    
    Instruction: \texttt{Given a sentence with an ``ambigious'' pronoun, either determine whether the sentence is inherently ambiguous (i.e., the thing that the pronoun refers to cannot be inferred by given information) or, if the pronoun can be implicitly deduced, state the antecedent of the pronoun (i.e., the noun to which the pronoun refers).}

    Input: \texttt{In the following sentences, explain the antecedent of the pronoun (which thing the pronoun refers to), or state that it is ambiguous.}\\
    \texttt{Sentence: The patient was referred to the specialist because he had a rare skin condition.}
    
    \texttt{Options:}
    
    \texttt{(A) The patient had a skin condition}\\
    \texttt{(B) The specialist had a skin condition}\\
    \texttt{(C) Ambiguous}

    Output: \texttt{(A)}
    
    \item \textbf{Dyck Languages.} 
    
    Instruction: \texttt{Predict the sequence of the closing parentheses of a Dyck-4 word without its last few closing parentheses.}

    Input: \texttt{Complete the rest of the sequence, making sure that the parentheses are closed properly. Input: [ [}

    Output:  \texttt{] ]}
\end{enumerate}

\subsection{WikiLingua Examples}
\label{wikilingua_examples}
Here are examples for WikiLingua summarization task.
\begin{enumerate}[noitemsep,topsep=2pt,leftmargin=*]
    \item \textbf{English-to-French summarization}
    
     Instruction: \texttt{Summarize the following English paragraph and then translate the English summary into a French summary}

     Input: \texttt{Look online at your town or city's codes, by-laws or dog legislation. There may be a code against unruly pets or incessant barking at night; many places have legislation or regulations in place that deals specifically with dogs and/or noise. There might also be a code covering ignoring requests from neighbors.   Often neighborhood or civil dispute centers produce small briefs on dog issues, as they're rather commonplace complaints. See if a precedent has already been set in your neighborhood. You may want to share your findings with your neighbor to give him or her one last chance to change before you call the authorities. If you're pretty sure it won't work, move straight to the next step. Find out what town hall/council/municipal office or other relevant authority to call so you can file a report on your neighbors for a noise complaint.The authorities will talk to the dog owner and assess the situation. They will usually inform you of the outcome.....}

     English summary: \texttt{Research your town or city's anti-barking laws. Call the relevant authority to report a noise complaint. Call animal control to report abuse. Get other neighbors to file the same complaint. Sue the dog owner in small claims court.}
    
     French summary: \texttt{Faites des recherches sur les lois contre les aboiements de chiens dans votre ville ou localité. Faites appel à l’autorité compétente pour enregistrer votre plainte pour nuisance sonore. Appelez le service de contrôle des animaux pour leur faire part de l’abus. Persuadez d’autres voisins de déposer la même plainte. Poursuivez le propriétaire de l’animal en justice.}
     
    \item \textbf{Spanish-to-English summarization}

    Instruction: \texttt{Summarize the following Spanish paragraph and then translate the Spanish summary into an English summary}

     Input: \texttt{Si bien no deseas gritar, debes hablar lo bastante alto para que las personas no tengan que pedirte que repitas lo que dices. Hablar en voz baja hará que las personas piensen que eres tímido, que no tienes confianza en lo que dices y que en realidad no quieres ser oído de todos modos.  Si hablas en voz baja, no tan solo los demás no podrán escuchar lo que dices, sino que también harás el papel de una persona con actitud sumisa, lo que insinúa lo contrario a una persona con confianza. Por otra parte, no querrás hablar tan alto que parezca que les estés hablando a las personas solo para que te escuchen. Tus palabras deben atraer por sí solas la atención del público. Lee tanto como puedas, desde revistas en línea (como Slate or Salon.com) hasta obra literarias serias (como Ana Karénina). Sabrás más mientras más leas, del mismo modo, tu vocabulario también se expandirá. Sin siquiera saberlo, aprenderás nuevas palabras y comprenderás nuevas frases y pronto estarás utilizando las palabras que leíste mientras estés hablando. Si realmente quieres hablar correctamente, tener un vocabulario amplio es una necesidad.  Esto no significa que cada vez debas utilizar cincuenta palabras de cien dólares en tus discursos o conversaciones diarias...}

     Spanish summary: \texttt{Habla con una voz lo bastante alta para que todos puedan oírte. Expande tu vocabulario. Evita usar demasiadas jergas. No tengas miedo de detenerte. Utiliza gestos solo cuando sea necesario. Sé más conciso. Repite tus ideas principales. Utiliza ejemplos concretos para ganarte a tu audiencia.}

     English summary: \texttt{Speak loudly enough for everyone to hear. Expand your vocabulary. Avoid excessive slang. Don't be afraid to pause. Use gestures only when necessary. Be more concise. Repeat your main points. Use concrete examples to grab your audience.}
\end{enumerate}
\subsection{Data Generation Details}
On average, a task in \textsc{Sup-NatIns} has 3,106 instances. An example of a \texttt{$<$instruction, input, output$>$} tuple is as follows: \texttt{$<$Translate the following English sentence into French, the food is good, la nourriture est bonne$>$}. The same instruction can correspond multiple input-output pairs. Since \textsc{Sup-NatIns} contains tasks in 55 different languages, we only select tasks with English as its input language or the output language to make sure the chain is connected. As a result, we only use 1,341 unique tasks out of 1,616 tasks from \textsc{Sup-NatIns}. For our single-task instruction tuning data, we randomly sample 10 combinations for each input-output pair, resulting in 13,410 instances. 

In this step, we random sample one task from each category, resulting in 3,024 pairs of two single instructions which are going to undergo the composability check. After composability check by LLM, we have 1,115 composable instruction pairs.

We later expand the dataset. For each valid category in the first subtask instruction for a compositional instruction with $\sigma = 2$, we randomly sample at most two more subtasks.  Consequently, each category is associated with a maximum of three distinct tasks, with some categories having only one task. Then in total, we have 2,933 training instances and 588 test instances for 2-instruction problem and 2,187 training instances and 480 test instances for the 3-instruction problem. 
We also have another variations of training set where we mix the training instances with 1 and 2 instructions (CoI$_{12}$) and with 1, 2, and 3 instructions (CoI$_{123}$). 

\begin{table*}[]
\small
\centering
    \begin{tabular}{l l}
    \toprule
    \rowcolor{lightgray}
    \multicolumn{2}{l}{\textbf{Single Instruction} from \textsc{Sup-NatIns} \cite{wang-etal-2022-super}}\\ \midrule
    \textbf{Task category:} & Generation \\
    \textbf{Original instruction:} & You are given a paragraph about various topics. Your task is to generate a title for the given paragraph.\\
    \textbf{Simplified instruction:} & Generate a title \\
    \textbf{Input:} & \textit{Ratatouille is a traditional stew made with summer vegetables. When cooking Ratatouille, you can...}\\
    \textbf{Output:} & Ratatouille: A Taste of Summer's farewell \\
    \hline
    \textbf{Task category:} & Style Transfer \\
    \textbf{Original instruction:} &  Convert the style of the given text to be suitable for a blog. Please consider the following points...\\
    \textbf{Simplified instruction:} & Convert the text into a blog-style text \\
    \textbf{Input:} & \textit{You can put an arrow on an elbow connector if you use the app.}\\
    \textbf{Output:} & Hey, there! Ever wondered if you can spice up your elbow connectors with a sleek arrow?\\
    \midrule
    \rowcolor{lightgray}
    \multicolumn{2}{l}{\textbf{Compositional Instruction} (Ours)} \\ 
    \midrule
    \textbf{Input:} & \textit{Ratatouille is a traditional stew made with summer vegetables. When cooking Ratatouille, you can...}\\
    \midrule
    \textbf{CoI variant:} & CoI \\
    \textbf{Instruction:}  & \textcolor{red}{Generate a title} and then \textcolor{darkgreen}{convert the title into a blog-style title} \\
    \textbf{Output:} & \textcolor{red}{Ratatouille: A Taste of Summer's Farewell} \\
    & \textcolor{darkgreen}{Delight in Ratatouille: Savor the Flavors of a Summer Farewell}
    \\
    \hline 
    \textbf{CoI variant}: & Concise CoI (C-CoI) \\
    \textbf{Instruction:} & Given a paragraph, \textcolor{gray}{write a blog-style title}\\
    \textbf{Output:} & \textcolor{red}{Ratatouille: A Taste of Summer's Farewell} \\
    & \textcolor{darkgreen}{Delight in Ratatouille: Savor the Flavors of a Summer Farewell}
    \\
    \hline 
    \textbf{CoI variant}: & Irrelevant CoI \\
    \textbf{Instruction:}&  \textcolor{red}{Generate a title} and then \textcolor{darkgreen}{convert the title into a blog-style title} \\
    \textbf{Output:} & \textcolor{red}{Ratatouille: A Taste of Summer's Farewell} \\
    &\textcolor{darkgreen}{Hey, there! Ever wondered if you can spice up your elbow connectors with a sleek arrow?} \\
    \bottomrule
    \end{tabular}
    \caption{
    Single instruction examples (top) and compositional instruction example in with CoI variants (bottom).
    The first subtask is to ``generate a title'' and the second subtask is to ``convert the text into a blog-style text.'' \textcolor{red}{Red text} refers to the first subtask, \textcolor{darkgreen}{green text} for the second subtask, \textcolor{gray}{gray text} for the concise compositional instruction. The input for the first task in the compositional instruction is the same, which is a text about Ratatouille.
}
    \label{tab:example_data}
\end{table*}

Specifically, we randomly select one task per category from \textsc{Sup-NatIns}, resulting in 76 task instructions with their input and output. 
Then we apply some heuristics (e.g., classification problems cannot be an input for the second or third instruction), resulting in 3,024 possible pairs of first and second instructions. 


\subsection{Heuristics for Validity Check}
\label{appendix:heuristics}
A list of tasks that cannot be task 1 (and task 2 if $\sigma=3$): Word Relation Classification, Preposition Prediction, Word Semantics, Entity Generation, Linguistic Probing, Textual Entailment, Misc., Sentence Perturbation, Text Categorization, Toxic Language Detection, Sentiment Analysis, Commonsense Classification, Language Identification, Stereotype Detection, Grammar Error Detection, Text Quality Evaluation, Irony Detection, Spam Classification, Punctuation Error Detection, Coherence Classification, Ethics Classification, Cause Effect Classification, Dialogue Act Recognition, Sentence Ordering, Discourse Relation Classification, Question Understanding, Discourse Connective Identification, Speaker Identification, Section Classification, Linguistic Probing, Dialogue State Tracking, Answerability Classification, Coherence Classification, Paper Review, Answer Verification, Entity Relation Classification, Speaker Relation Classification, Stance Detection, Fact Verification, Text Matching, Intent Identification, Word Relation Classification.

\begin{table*}[]
    \centering
    \small
    \begin{tabular}{l| r r | r r | r r | r r | r r}
    \toprule
    \textbf{Model} & \multicolumn{10}{c}{\textsc{Rouge-L}} \\ \midrule
  Baseline & \multicolumn{2}{c|}{CoI$_2$-test}  & \multicolumn{2}{c|}{C-CoI$_2$-test}  & \multicolumn{2}{c|}{CoI$_3$-test} & \multicolumn{2}{c}{C-CoI$_3$-test} &  \multicolumn{2}{c}{BBH}\\
  & M & A & M & A & M & A & M & A & M & A\\ \midrule
    Base &  24.93 &  24.95 & 23.09 & 25.12  & 23.66 & 20.99 & 21.55 & 22.02 & 8.51 & 14.36 \\
    CoT &  16.61 &23.82 & 17.41 & 24.42 & 16.90 &20.09  & 17.83 & 21.35 &  5.84 & 17.05\\
    CoI$_1$ & 39.72 & 32.32 & 36.30 & 28.53 &29.62& 21.75 & 27.70 & 22.63 & 27.68 & 28.74 \\
    CoI$_1$-long & 30.08 & 22.68 & 27.90 &  22.42& 21.87 & 17.82 & 21.45 &  18.31 & 24.01 &29.87
    \\
    \midrule
    \multicolumn{7}{l}{\textbf{Chain-of-Instructions Models}} \\
    \midrule
    CoI$_2$ & 60.43 & 62.04 & 57.65 & \textbf{56.59} & 48.31 & 48.23 & 48.67 & 46.97 & 10.65 & 12.11 \\
    CoI$_3$  & 33.63 & 31.62 & 32.02 & 29.54 & 60.03 & 47.03 & 56.77 & 44.36 & 5.78 & 7.00\\
    CoI$_{12}$ & \textbf{70.76} & \textbf{67.50} & 53.50 & 46.27 & 59.84  & 50.23 & 41.63 & 39.23 & 24.44 & 28.80 \\
    CoI$_{123}$  & 45.16 & 67.12 & 31.32 & 44.93 & 61.61 & \textbf{67.49} & 29.61 & 44.40 & \textbf{29.39} & 27.57\\ \midrule
    \multicolumn{5}{l}{\textbf{Concise \textsc{CoI} Models}} \\ \midrule
    C-CoI$_2$ & 59.07 & 34.59 & 58.33 & 42.60 & 51.34 & 32.75  & 49.77  & 39.43 &7.95  &12.25\\
    C-CoI$_3$  & 33.63  & 31.28 & 32.02 & 29.54 & 60.03 &47.38  &56.77 &44.36 & 5.79 & 7.00\\
    C-CoI$_{12}$ & 59.19 & 47.36 & 56.94 & 47.70 & 51.07 & 42.99  & 46.42 & 38.16 & 28.84 & \textbf{31.26}\\
    C-CoI$_{123}$  & 69.47 & 54.74 & \textbf{66.17} & 51.94 & \textbf{68.97} & 58.85  & \textbf{58.15} & \textbf{49.60} & 24.04 & 28.91 \\ \midrule
    \multicolumn{5}{l}{\textbf{Irrelevant \textsc{CoI} Models}} \\ \midrule
    I-CoI$_2$ &  45.09 & 48.56 & 44.57 & 47.03 & 36.02 &38.02  & 35.96 & 38.24 & 10.25 & 21.37\\
    I-CoI$_3$  & 33.73  & 28.64 &  31.91 & 27.27 & 34.16& 32.02  & 33.67 & 30.42 & 6.31 & 6.21\\
    I-CoI$_{12}$ & 53.53 & 50.05 & 42.33	& 37.98 & 39.42 &38.24  & 31.98 & 32.10 & 23.91 & 27.27\\  
    I-CoI$_{123}$  & 54.59& 50.60 & 38.91  & 36.27& 37.26& 36.08  & 26.93 & 33.59 & 28.95 & 28.65\\
    \bottomrule
    \end{tabular}
    \caption{Results on compositional instruction test sets and BIG-Bench Hard (BBH). CoT refers to chain-of-thought prompting with the non-fine-tuned base model. M refers to Mistral 7B-Instruct and A is Alpaca-7B. CoI$_1$-long refers to the same set of CoI$_1$ but the instructions are not summarized.}
    \label{tab:full_results}
\end{table*}
\subsection{Downstream Task}
To find generated summaries in source and target languages, for the CoI-tuned models, we use keyword matching and text processing (e.g., ``Task 1 input and task 2 output.'', ``Task 2 output'', ``1 output'', ``2 output'').  Finding summaries in source and target languages from the baseline model is trickier because there is no clear separation between the sentences in the generated output. Thus, we split the first output and the second output from the baseline is by using a language identification model.\footnote{\url{https://pypi.org/project/langdetect/}} If the detected language is the same with the source language, we label the output as the source language summary. If the detected language is the same with the target language, we label the output as the target language summary. 

\begin{table}[]
    \small
    \centering
    \addtolength{\tabcolsep}{-0.3em}
    \begin{tabular}{@{}l|  r r | r r |@{}}
    \toprule
    \multirow{2}{*}[-1ex]{Model}& \multicolumn{2}{c|}{CoI$_4$-test}  & \multicolumn{2}{c|}{CoI$_5$-test}\\  \cmidrule{2-5}
   & M & A & M & A \\ \midrule
    Base &  22.03 & 17.48  & 20.95 &  15.79\\
    CoI$_1$ & 19.18 & 16.44 &  15.16 &  12.87 \\ 
    CoI$_{12}$ & 47.36 & 40.59 & 37.04 & 33.05 \\
    CoI$_{123}$  & 55.24 & 53.33 & 46.43 & 44.87\\
    \bottomrule
    \end{tabular}
    \caption{\textsc{Rouge-L} results on CoI-4,5 test sets. 
    }
    \label{tab:coi45test_experiment}
    \vspace{-5mm}
\end{table}

\subsection{Full results}
\label{appendix:fullresults}
\paragraph{Another variant of CoI: Concise CoI} 
We have another variant of CoI with shorter compositional instructions and we call this variant as Concise-Co, or C-CoI, (see Table \ref{tab:example_data}). The purpose of training a model with C-CoI dataset is to examine whether shorter compositional instructions can perform better. To create this dataset, we perform a 2-shot prompting on LLM with the following prompt: ``\textit{Summarize the input to a single coherent sentence which contains only 20 words or fewer without changing the meaning.}'' The input and output for this C-CoI training instance is the same as those for CoI. 

We have another variant of CoI which we call as irrelevant CoI as an ablation study. For the irrelevant CoI, we use the same instruction and input as CoI, but we assign a random output from the original benchmarking single-instruction dataset for the second task rather than the correct generated output. 
Thus, the second task's output is not relevant with the first task's output. Using the examples inin Table \ref{tab:example_data}, for the irrelevant CoI, the second output is not the blog-style title about Ratatouille but rather a \textit{blog-style text about adding an arrow to an elbow connector in slides}. We try this irrelevant CoI as another variant to see whether the correctness of the output from the second task or third task actually matters for the model improvement. Our finding is that the correctness for the second and third outputs is important for compositional test sets even though it does not affect much for single task problems as shown Table \ref{tab:full_results}. 

\paragraph{Does concise CoI instruction help improve performance?} \label{sec:concise_coi_results}
We also show the scores of Concise-CoI in Table~\ref{tab:multi_instruction_experiment}. For C-CoI models, we notice that C-CoI$_{123}$ Mistral consistently performs better than CoI models for all datasets except for BBH. Meanwhile, for C-CoI Alpaca, the result is quite mixed, but interestingly both C-CoI$_{12}$ and C-CoI$_{123}$ Alpacas perform better than CoI$_{12}$ and CoI$_{123}$ on BBH, suggesting that shorter instructions help more on unseen single tasks.

\begin{figure*}
    \centering
    \includegraphics[width=\linewidth]{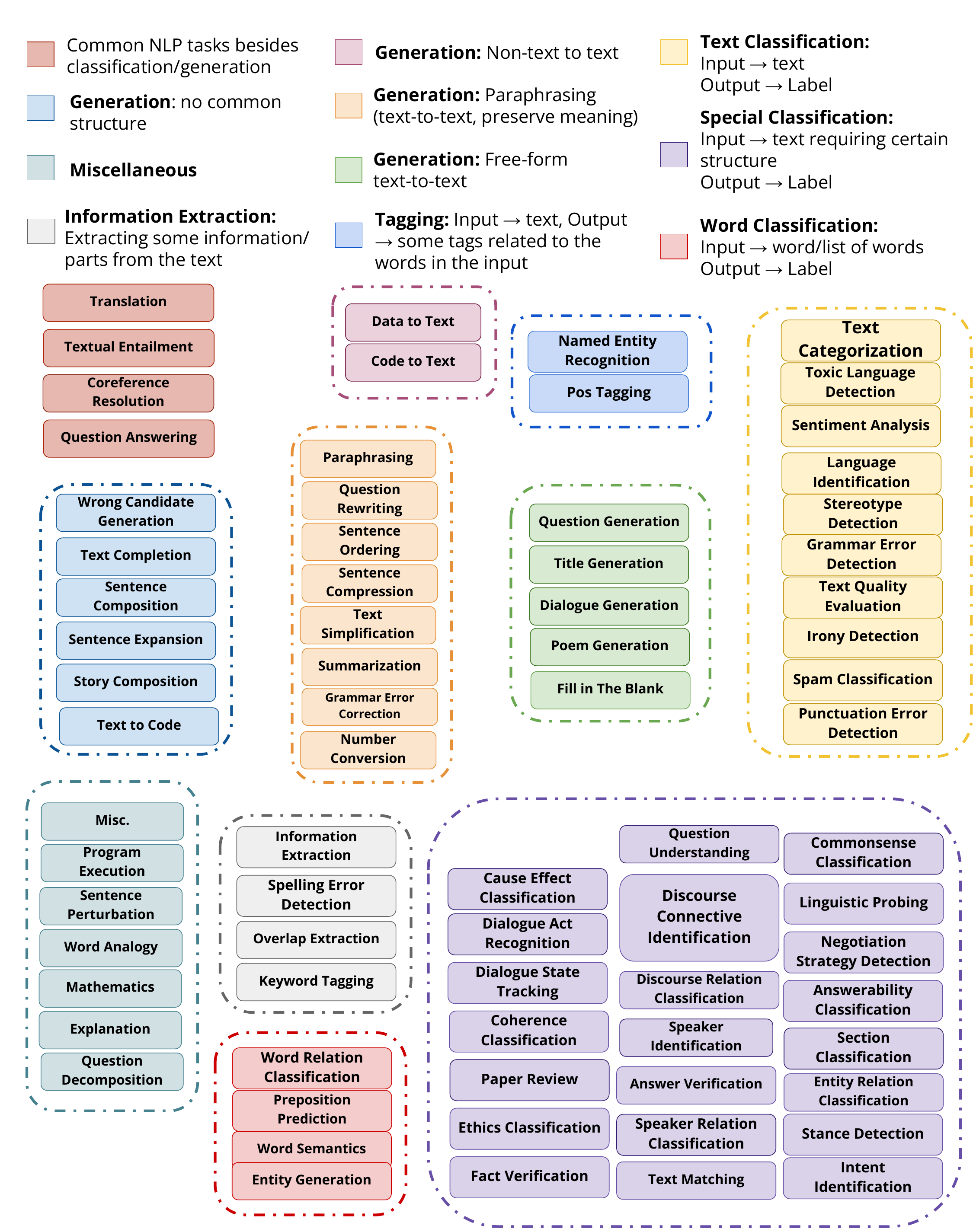}
    \caption{We manually classify task categories from \textsc{Sup-NatIns} into eleven groups based on their input and output types. This serves as a heuristic for selecting tasks that are composable through human reasoning.}
    \label{fig:task_categories}
\end{figure*}

\subsection{Full Intermediate Results}
Table \ref{tab:intermediate_result_valid} shows how many valid outputs for the first subtasks and second subtasks from baseline and CoI models. 

\begin{table}[t]
    \centering
    \small
    \begin{tabular}{l | r r | r r} 
    \hline
     \textbf{Metric} &   \multicolumn{2}{c|}{\textbf{Mistral}}  & \multicolumn{2}{c}{\textbf{Alpaca}} \\
     Rouge-L & Base & CoI & Base & CoI 
      \\ \hline
     \multicolumn{5}{c}{\textbf{Test Set: CoI$_2$}}
     \\
     \hline
        Subtask 1 &  1.32 & 90.50 &  13.98 & 84.16 
        \\
        Subtask 2 & 2.40 & 49.21 & 7.02 & 45.57\\
        \hline
        \#valid 1st output & 15 & 588 &  207 & 588\\ 
        \#valid 2nd output & 33 & 588 & 144 & 588 \\ 
    \hline
     \multicolumn{5}{c}{\textbf{Test Set: CoI$_3$}}
     \\ \hline
    Subtask 1 & 18.04 & 81.49 & 9.56 & 91.77\\
    Subtask 2 & 6.82 & 68.65 &  2.13& 71.67\\
    Subtask 3 & 6.93 & 32.73 &  3.30& 35.52
        \\ \hline
    \#valid 1st output &  259 & 479 &  120 & 480 \\ 
    \#valid 2nd output & 136 & 479 & 45 & 480 \\ 
    \#valid 3rd output & 185 & 463 & 81 & 478\\ 
    \hline
    \end{tabular}
    \caption{Results on intermediate tasks. CoI models refer to best models of CoI: CoI$_{12}$ if the test set=CoI$_2$ and CoI$_{123}$ if the test set=CoI$_3$}
    \label{tab:intermediate_result_valid}
    \vspace{-2mm}
\end{table}

\end{document}